\documentclass{llncs}
\usepackage{amssymb}
\usepackage{graphicx}
\usepackage{subfigure}
\usepackage[misc]{ifsym}
\usepackage{amsmath}
\usepackage{lineno}
\usepackage{subfig}
\usepackage{stmaryrd}
\usepackage{url}
\usepackage{float}
\urldef{\mailsa}\path{zxysee@bupt.edu.cn}
\urldef{\mailsb}\path|{yifan_liu|
\urldef{\mailsc}\path|zcqin}@buaa.edu.cn|
\urldef{\mailsd}\path{lijiahong@meituan.com}
\newcommand{\keywords}[1]{\par\addvspace\baselineskip
\noindent\keywordname\enspace\ignorespaces#1}

\title{Emotion Classification with Data Augmentation Using Generative Adversarial Networks}

\author{Xinyue Zhu$^{1,2}$  \and Yifan Liu$^{2}$ \and Zengchang Qin$^{*2}$ \and Jiahong Li$^{3}$}
\institute{$^1$School of Electronic Engineering\\
Bejing University of Posts and Telecommunications, Beijing 100876, China\\
$^2$Intelligent Computing and Machine Learning Lab, School of ASEE\\
Beihang University, Beijing 100191, China\\
$^3$Beijing San Kuai Yun Technology Co., Ltd.\\
Hengdian Building, No.4 Wangjing East RD, Chaoyang District, Beijing, China\\
$^1$\mailsa,\\
$^2$\mailsb,
\mailsc\\
$^3$\mailsd\\
}

\begin{document}
\maketitle

\begin{abstract}
It is a difficult task to classify images with multiple class labels using only a small number of labeled examples, especially when the label (class) distribution is imbalanced. Emotion classification is such an example of imbalanced label distribution, because some classes of emotions like \emph{disgusted} are relatively rare comparing to other labels like {\it happy or sad}.
In this paper, we propose a data augmentation method using generative adversarial networks (GAN). It can complement and complete the data manifold and find better margins between neighboring classes. Specifically, we design a framework using a CNN model as the classifier and a cycle-consistent adversarial networks (CycleGAN) as the generator. In order to avoid gradient vanishing problem, we employ the least-squared loss as adversarial loss. We also propose several evaluation methods on three benchmark datasets to validate GAN's performance. Empirical results show that we can obtain 5\%$\sim$10\% increase in the classification accuracy after employing the GAN-based data augmentation techniques.

\keywords{Data augmentation; Emotion classification; Imbalanced data processing; GAN; CycleGAN.}
\end{abstract}

\section{Introduction}
In recent development of deep learning, neural networks with more and more layers are proposed \cite{he2016deep}\cite{simonyan2014very}. Such neural network models have much larger capacity that needs a larger training set.
%Otherwise, training network with a large number of weights and variables would easily get over-fitting if insufficient training samples were provided.
However, it is always expensive to obtain adequate and balanced dataset with manual labels. This has been a general problem in machine learning as well as in computer vision.
An effective way of synthesizing images to supplement training set may help boost accuracy in image classification.
%and automatically obtain samples with specific features and given labels becomes a viable solution.
Using data augmentation for enlarging training set in image classification has reported in various literatures \cite{cirecsan2011high}\cite{krizhevsky2012imagenet}\cite{simard2003best}. Model performance can be improved because data augmentation can overcome the problem of inadequate data and imbalanced label distribution.
%However, they contribute little to supplement the data manifold since only imagelevel samples are generated in this process.
However, it is an unsolved problem of how to generate (sample) data from the `true distribution' of given limited training data.
In this research, based on Generative Adversarial Networks (GANs), we propose a new method for data augmentation
in order to generate new samples via adversarial training, thus to supplement the data manifold
to approximate the `true distribution' and that may lead to better margins between different categories of data.

Since the invention of GAN \cite{goodfellow2014generative}, it has been well used in different machine learning applications \cite{denton2015deep}\cite{yu2016seqgan}, especially in computer vision and image processing \cite{isola2016image}\cite{ledig2016photo}. In this paper, we explore how to use GAN to generate
images helping enlarge original dataset effectively and balance label distribution from data augmentation. We focus on the emotion (facial expression images) classification task because it is a typical classification task with inadequate data and imbalanced label distribution. On one hand, the training dataset obtained from the laboratory are limited to diversity and quantity. On the other hand, some classes of emotion images (such as \emph{disgust}) have few samples in the training set from the real world than other classes, also some images can be very nuanced, making it difficult even for humans to agree on their correct labeling \cite{ng2015deep}.
In our work, we build a classical convolutional neural network (CNN) classifier for emotion image classification and train the CycleGAN model \cite{zhu2017unpaired} with least-squared loss \cite{mao2016least} to achieve image-to-image transformation, which can synthesis images for the unusual classes of emotion from a related image source.
%Our aim is to explore the effect of data augmentation using GAN, so we build a relatively shallow CNN model rather than an extremely powerful one, which is only requested to extract general features for each class and have a certain ability to distinguish among them. Contrary to this,
Given a classical CNN classifier,
our effort is on constructing GAN model for improving performance in generating images given a specific class. The CycleGAN model is used for image translation between two unpaired domains.

 %to generate target images with insufficient samples from large reference samples.
%In our research, we first train a classifier using original samples as our baseline. After that, we select one or more classes as our to-be-generated classes. In order to take advantage of existing samples, we choose a class as our reference class which has a large sample size. After successfully training the CycleGAN model, we export the graph and add generated images to original dataset before retraining the classifier.

The main contributions of this paper can be summarized as follows.
%\begin{itemize}
%\item
(1) We propose a framework for data augmentation by using GAN to generate supplementary data in emotion classification task. (2) Through empirical studies on three benchmark datasets, we found that performance of the new model is significantly improved compared to the baselines.
%\item	
(3) We combine least-squared loss from LSGAN and adversarial loss in CycleGAN to avoid the problem of vanishing gradients, this is verified to be effective in training process.
%\item
%We analyze the problem of imbalanced distribution in classification task, and CycleGAN's positive role to resolve this problem.
%\item
% (3) We show that GAN-based augmentation model performs better than other augmentation methods. By possessing a more complete data manifold, the classifier can learn to find margins or hyper-planes between neighboring classes better. [the 3rd contribution is deleted and this one is not so good]
%\end{itemize}

\section{Related Work}
\subsection{Learning with Imbalanced Emotion Datasets}
Facial expression recognition (or emotion classification) has attracted much attention in computer vision in past few decades. Current techniques related to facial expression mainly focus on recognizing seven prototypical emotions (\emph{neutral}, \emph{happy}, \emph{surprised}, \emph{fear}, \emph{angry}, \emph{sad}, and \emph{disgusted}), which are considered basic and universal emotions for human. Such recognition is sometimes very difficult since there is only a slight difference between different emotions, which requires an efficient and subtle feature extractor to be trained. Moreover, Ng $et$ $al.$ \cite{ng2015deep} pointed out that the imbalanced distribution among emotion classes may lead to low accuracy in classes with fewer samples. To deal with imbalanced datasets, many methods were proposed, such as undersampling \cite{liu2009exploratory}, synthesizing minorities \cite{chawla2002smote}, creating `box' around minorities \cite{goh2014box} and etc. Different from these works, we aim to resolve this problem by generating data of minority classes from low-dimensional manifold, which improves the data distribution from feature level.

\subsection{Generative Adversarial Networks}
Generative adversarial networks (GANs) can be used to generate images from an adversarial training. The generator attempts to produce a realistic image to fool the discriminator, which tries to distinguish whether its input image is from the training set or the generated set. Since the invention of GAN \cite{goodfellow2014generative} in 2014, variant models based on GAN were proposed \cite{dumoulin2016adversarially}\cite{mao2016least}\cite{mirza2014conditional}\cite{radford2015unsupervised}\cite{zhu2017unpaired}. Generative adversarial nets are now widely used in many image tasks such as single image super-resolution \cite{ledig2016photo}, image manipulation \cite{zhu2016generative}, synthesis \cite{denton2015deep} and image-to-image translation \cite{isola2016image}.

Zhu $et$ $al.$ \cite{zhu2017unpaired} proposed the CycleGAN. It can do image-to-image transition between two unpaired image domain, which is helpful to build our framework. The main idea of the cycleGAN is ``If we translate from one domain to another and back again we must arrive where we start" \cite{zhu2017unpaired}. The LSGAN \cite{mao2016least} used a least square distance to evaluate the difference between the distribution, which is more stable than the Jensen-Shannon (JS) divergence used in \cite{goodfellow2014generative} and convergence more quickly than the Wasserstein loss \cite{arjovsky2017wasserstein}.
In our research, we choose CycleGAN \cite{zhu2017unpaired} and the techniques in LSGAN \cite{mao2016least} to generate labeled emotion images and show that these images are helpful in final image classification task.
\subsection{Data Augmentation}
In the field of deep learning, where the scale of dataset has a great influence on the final outcome, data augmentation is often used to expand the training corpus. As for the existing techniques of data augmentation, they can be grouped into two main types: (a) geometric transformation which is relatively generic and computationally cheap and (b) task-specific or guided-augmentation methods which are able to generate synthetic samples given specific labels \cite{dixit2016aga}. In the case of image classification, the first group of data augmentation methods always focus on generating image data through label-preserving linear transformations (translation, rotation, scaling, horizontal shearing) such as Affine \cite{cirecsan2011high}, elastic deformations \cite{simard2003best}, patches extraction and RGB channels intensities alteration \cite{krizhevsky2012imagenet}. However, if we look deeper into these methods, they only lead to an image-level transformation through depth and scale and actually not helpful for dividing a clear boundary between data manifolds. Such data augmentation does not improve data distribution which is determined by higher-level features. For the second group, more complex manually-specified augmentation schemes are proposed. For instance,  authors in \cite{hauberg2016dreaming} proposed an approach to learn multivariate normal distribution of each class in the whole mean manifold. In \cite{dixit2016aga}, an attribute-guided augmentation in feature space is designed. In the field of 3D motion capture, 2D images are used for generating 3D ones \cite{rogez2016mocap}. Our approach aims to solve similar task in \cite{hauberg2016dreaming} but is very different from all these methods above. In this paper, new training corpus is generated from CycleGAN, which remain high-level features extracted from original images.

\section{Data Augmentation Using CycleGAN}
\begin{figure}[htb]
\begin{centering}
\setlength{\abovecaptionskip}{0.cm}
\setlength{\belowcaptionskip}{-0.cm}
\includegraphics[width=1\textwidth]{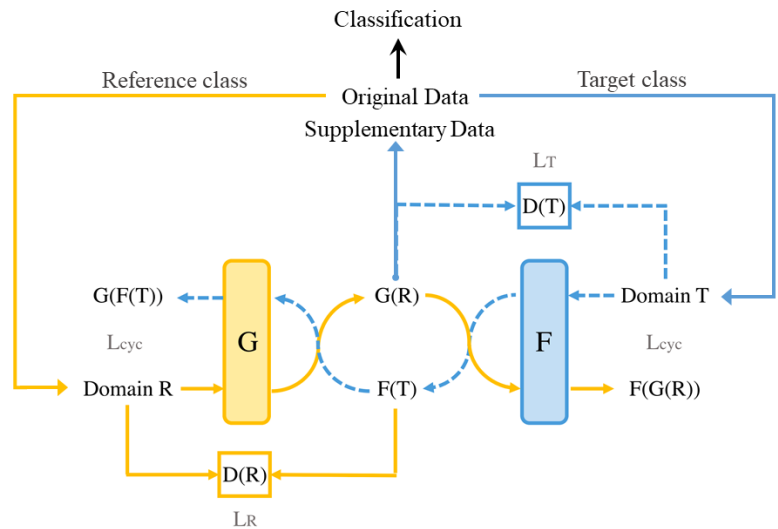}
\par\end{centering}

\caption{\label{fig:pp} An illustration of the proposed framework of using CycleGAN for data augmentation and classification using a CNN classifier.}
\end{figure}

Fig. \ref{fig:pp} shows our framework of GAN-based data augmentation. Both reference images and target images are collected from the original data and flow into the CycleGAN as domains $R$ and $T$, respectively. $G$ and $F$ are two generators, transferring $R \to T$ and $T \to R$, respectively. Supplementary data is generated through generator $G$. A CNN classifier is trained using original data and supplementary data as input. In CycleGAN model, $L_{R}$ is the LSGAN loss relative to reference domains and $L_{T}$ is the LSGAN loss relative to target domains. Besides, a cycle loss, namely $L_{cyc}$, is calculated to keep cycle consistency of the whole model.

\subsection{Cycle-consistent Adversarial Networks}
In this work, CycleGAN \cite{zhu2017unpaired} is used to realize unpaired image-to-image translation, learning mapping functions between images of reference class ($R$) and target class ($T$). We use generators $G$ and $F$ to achieve domain transfer $G$: $R \to T$ and $F$: $T \to R$. Discriminators are denoted by $D_R$ and $D_T$, where $D_R$ aims to distinguish between real images in $R$ and translated fake images ${{F(T)}}$ in reference domain. $D_T$ is the discriminator in the
target domain. We not only want to make the generated images $G(R)$ look like the target images in $T$, but also the reconstructed images $F(G(R)) \approx R$ to guarantee the cycle-consistency in addition to the adversarial loss.
As for adversarial loss, $G$ tries to generate $G(r)$ which is so similar to $t$ that can fool the discriminator $D_T$. Therefore, the loss related to $G$ and $D_T$ is:
\begin{equation}
L(G, {D_T}, R, T) = {E_{t\sim{p_{data}}(t)}}[\log{D_T}(t)]
+{E_{r\sim{p_{data}}(r)}}[\log(1-{D_T}(G(r))]
\end{equation}
However, this logarithm form makes training and convergence difficult since it is likely to cause gradient vanishing problem \cite{arjovsky2017wasserstein}. Here we apply a least-squared loss proposed in LSGAN \cite{mao2016least} to avoid this phenomenon and maintain the same function as adversarial loss in original CycleGAN. For the reference domain $R$, the loss is defined by:
\begin{equation}
{L_{LSGAN}}(G, {D_R}, T, R)= {E_{r\sim{p_{data}}(r)}}[({D_R}(r)-1)^2]
+ {E_{t\sim{p_{data}}(t)}}[{D_R}(G(t))^2]
\end{equation}
For the target domain $T$, the loss is:
\begin{equation}
{L_{LSGAN}}(G, {D_T}, R, T)= {E_{t\sim{p_{data}}(t)}}[({D_T}(t)-1)^2]
+ {E_{r\sim{p_{data}}(r)}}[{D_T}(G(r))^2]
\end{equation}
We can then define the final loss by:
\begin{equation}
\begin{split}
L(G, F, {D_S}, {D_R}) &= {L_R} + {L_T} + {L_{cyc}}\\
&={L_{LSGAN}}(G, {D_R}, S, R) + {L_{LSGAN}}(F, {D_S}, R, S) + \lambda{L_{cyc}}(G, F)
\end{split}
\end{equation}
where cycle consistency loss ($L_{cyc}$) is defined by:
\begin{equation}
\begin{split}
{L_{cyc}(G, F)} &= {E_{r\sim{p_{data}}(r)}}[\|F(G(r)) - r\|_1]\\
&+{E_{t\sim{p_{data}}(t)}}[\|G(F(t)) - t\|_1]
\end{split}
\end{equation}
where $||\cdot||_1$ is the $L_1$ norm. With these loss functions, the final functions we aim to solve is:
\begin{equation}
G^*, F^* = arg \min_{F,G}\max_{D_T, D_R}L(G, F, D_T, D_R)
\end{equation}
other details of CycleGAN can be referred to \cite{zhu2017unpaired}.

\subsection{Class Imbalance and Data Manifold}
\begin{figure}
\centering
\begin{minipage}[b]{0.33\linewidth}
\centering
\includegraphics[width=1.5in]{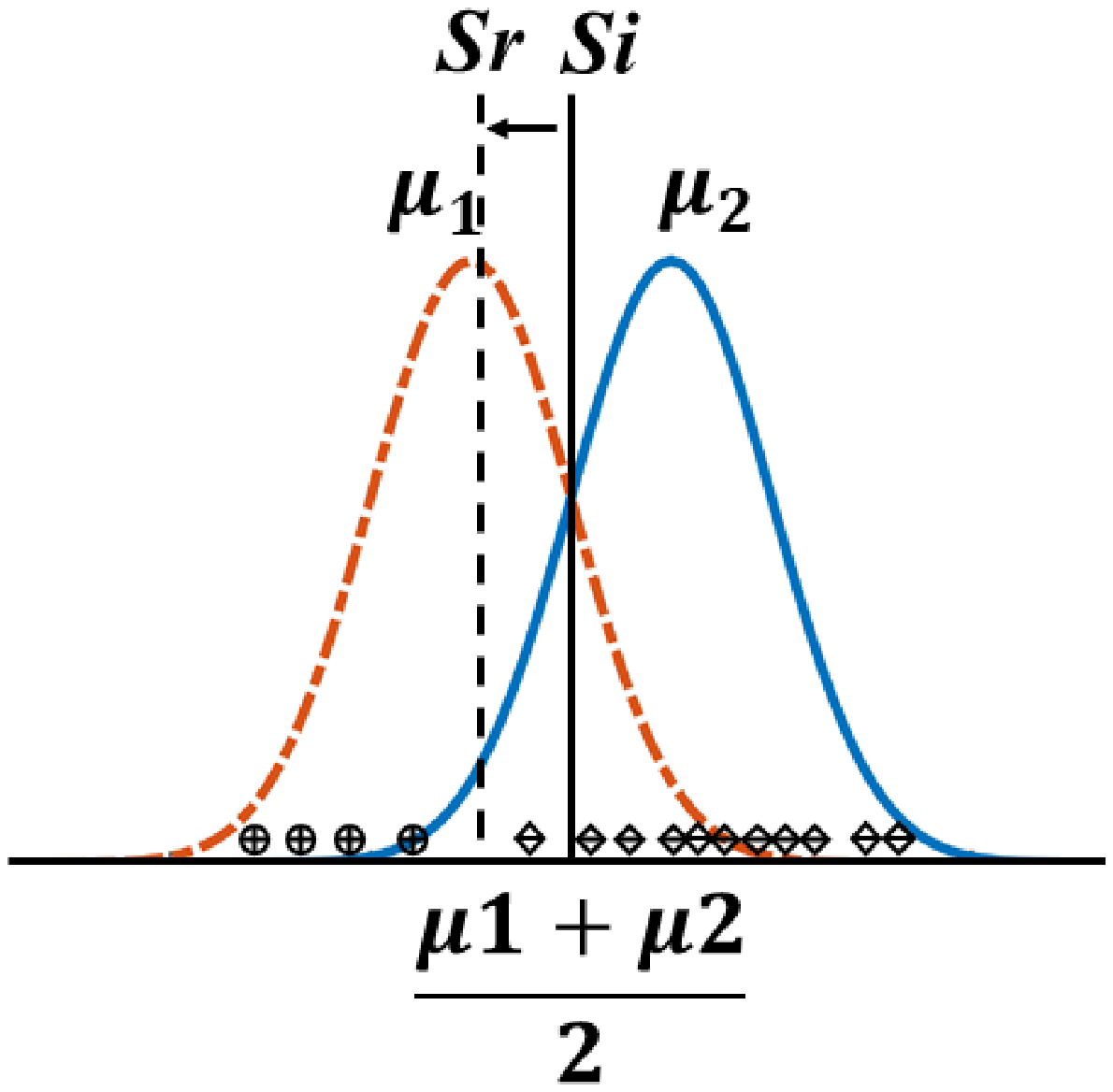}
\caption*{\label{fig:binary}(a)}
\end{minipage}%
\begin{minipage}[b]{0.33\linewidth}
\centering
\includegraphics[width=1.5in]{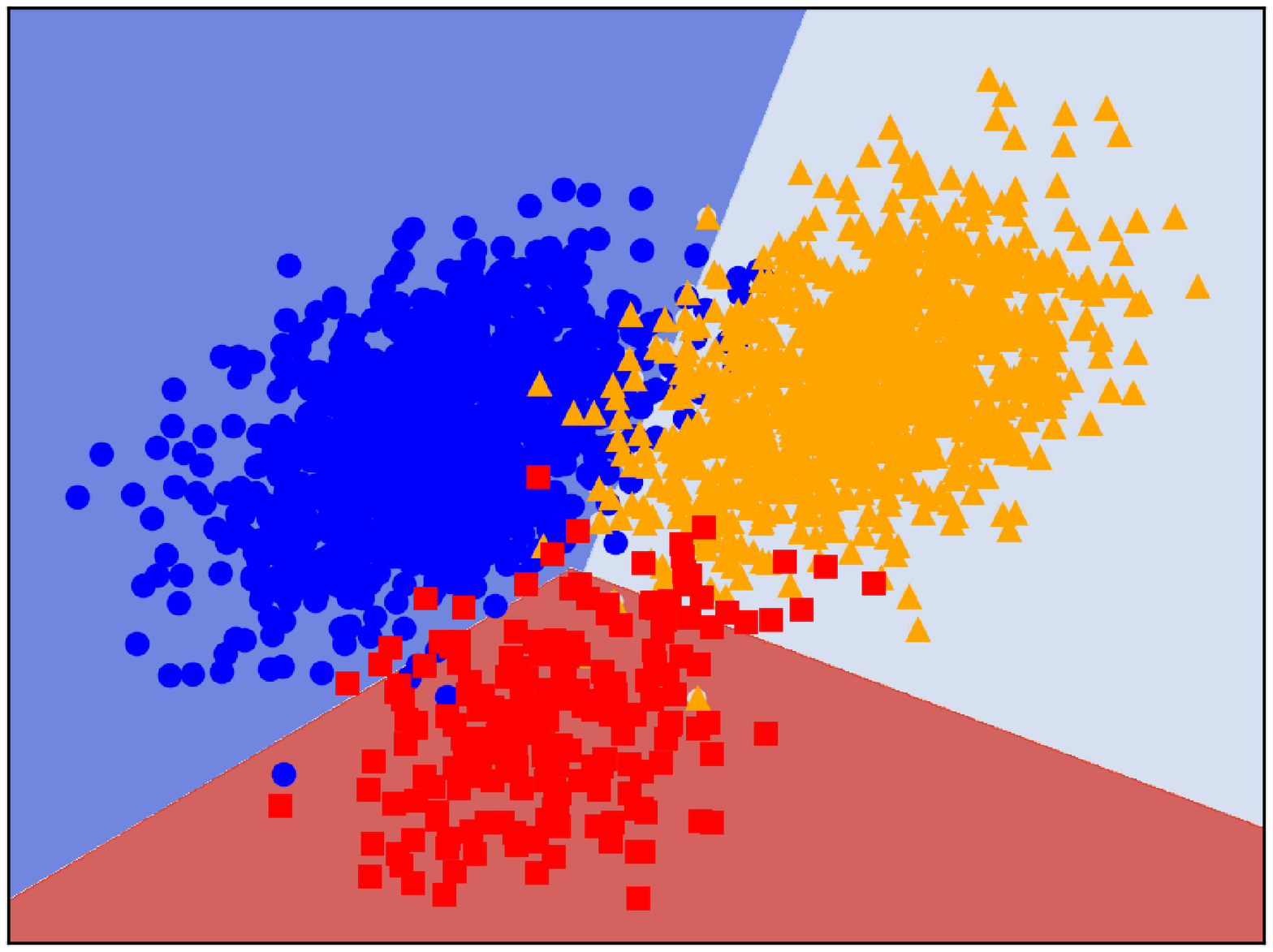}
\caption*{\label{fig:toy1}(b)}
\end{minipage}
\begin{minipage}[b]{0.33\linewidth}
\centering
\includegraphics[width=1.5in]{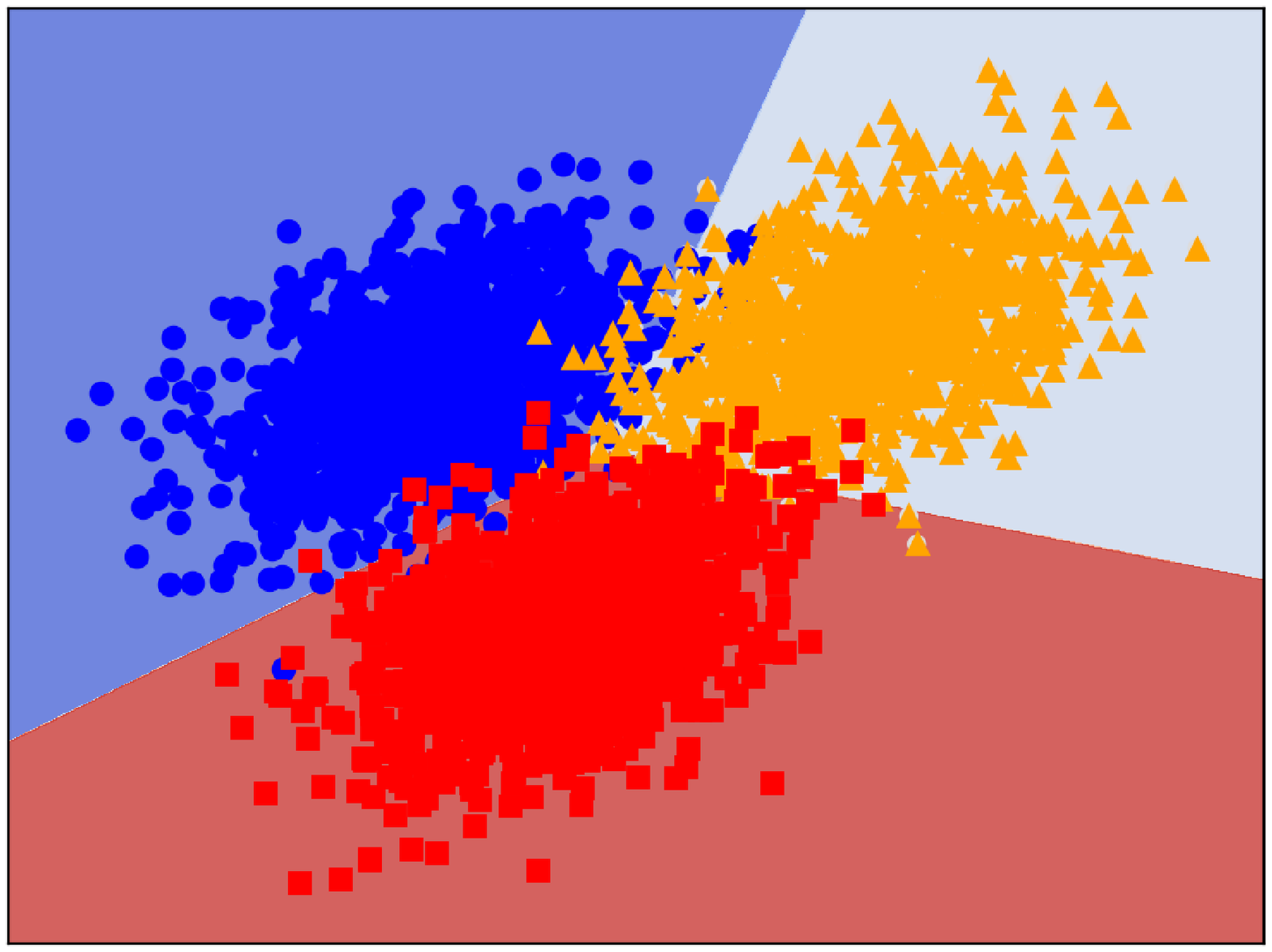}
\caption*{\label{fig:toy2}(c)}
\end{minipage}
\caption{\label{fig:dm} (a) A binary classification problem with one-dimensional data. (b) Data manifolds with learned boundaries with imbalanced distributions. (c) The manifolds of balanced distribution. Data are generated with 2-dimensional Gaussians and using SVMs with linear kernel as classifier.}
\end{figure}

When the classes have imbalanced distributions, the classifier prone to learn biased boundary between classes. Take an example of binary classification in one-dimensional sample. In Fig.\ref{fig:dm}-(a),  Class 1 and 2 are both generated from Gaussian distributions with the same standard deviation 1, and has the means of $\mu_1$ and $\mu_2$, respectively. Ideally, the boundary function x=($\mu_1$+$\mu_2$)/2, denoted by $S_i$ can distinguish between these two classes. However, an imbalanced distribution in two classes will result in a biased linear boundary $S_r$ moving towards the minor class, since given samples are insufficient to form a correct margin with minimized loss. Some detailed discussions can be referred to \cite{wallace2011class}.

Now back to our emotion classification task. Under the assumption that image samples lie on several sub-manifolds in a high-dimensional space, where images of the same emotion lie in the same sub-manifold, image classification task is actually a task to explore the underlying geometric structure of data distribution, thus to find best-split hyper-planes between different categories in this space. These hyper-planes divide the space into several parts according to margins, each represents a clustering of a specific class (Fig.\ref{fig:dm}-(c)).
When the dataset is imbalanced, it is very likely to form an incomplete manifold. Since in the same space, minorities are distributed more sparsely in their regions. In this case, biased margins or hyper-planes are learned, making it a difficult task for classifier to predict correct labels for given instances (Fig.\ref{fig:dm}-(b)). Although some data augmentation techniques mentioned in Section 2.3 can alleviate this problem from several aspects, the most essential solution is to further complement and complete the data manifold.

The reason we choose cycleGAN in stead of the classical GAN is  because the original GAN \cite{goodfellow2014generative} learns a mapping from low-dimensional manifold (directly determined by noise $z$) to high-dimensional data spaces, while the CycleGAN, as a tool for translation between two domains of both high-dimensional data, need to learn a low-dimensional manifold and also the parameters to map it back to high-dimensional space. Here we use $\mathbb{P}_{r}$ and $\mathbb{P}_{t}$ to represent real distributions of domain $R$ and $T$, respectively, and $\mathbb{M}$ is the low-dimensional manifold. As domain $R$ has only a small number of training samples, when it is projected into a low dimension space, sparse distribution cannot form a complete $\mathbb{M}$ with efficient feature information. By using CycleGAN, sufficient samples in $\mathbb{P}_{r}$ may lead to a more complete $\mathbb{M}$. $G_\theta$ can be learned through minimizing distance between $G(r)$ and $T$ (e.g, $||$G(r)$ - $T$||^2$) to 'pull' $\mathbb{M}$ to $\mathbb{P}_{t}$. If we again project $\mathbb{P}_{t}$ into manifold $\mathbb{M}$, they will form a meaningful feature-level manifold thanks to the generated samples.

\section{Experimental Studies}
\subsection{Toy Data Experiment}
Before doing experiments on emotion datasets, we first validate GAN's role in completing data manifold on a toy dataset. We use three two-dimensional Gaussian distributions ($x_1$, $x_2$) $\sim N_m$($\mu_m$, $\sigma_m$), $m\in \{1,2,3\}$ to simulate the distribution of three classes of data, where $\mu_1$=[0, 6], $\mu_2$=[6.5, 7], $\mu_3$=[2, 2] and the covariance matrix is ([2, 1]; [1, 2]) for all three distributions. Imbalanced dataset is artificially created by randomly sampling 1000, 1000 and 100 points from each class for training, and for each class, we have 100 data points for test. Support vector machines (SVM) with linear kernel function \cite{keerthi2003asymptotic} is employed for classification task. After that, we train a CycleGAN to generate 900 target class (minority class) from reference class (one of majority classes) and these supplementary samples are added to the original dataset and trained on the same SVM classifier.

We draw two figures (Fig.\ref{fig:dm}-(b) and (c)) to show the data distribution and learned margins before and after adding the CycleGAN-based augmentation. The original biased margins in imbalanced dataset Fig.\ref{fig:dm}-(b) show a clear change to more correct ones in Fig.\ref{fig:dm}-(b). Moreover, the classification accuracy has increased from 93.3\% to 98.0\%. Although the distribution and dimension of data in this toy experiment is much simpler than real image data, the results can validate the effectiveness of the new approach by improving data manifold in imbalanced datasets by doing data augmentation.

\subsection{Benchmark Datasets}
\begin{figure}[htb]
\begin{centering}
\begin{centering}
\includegraphics[width=1\textwidth]{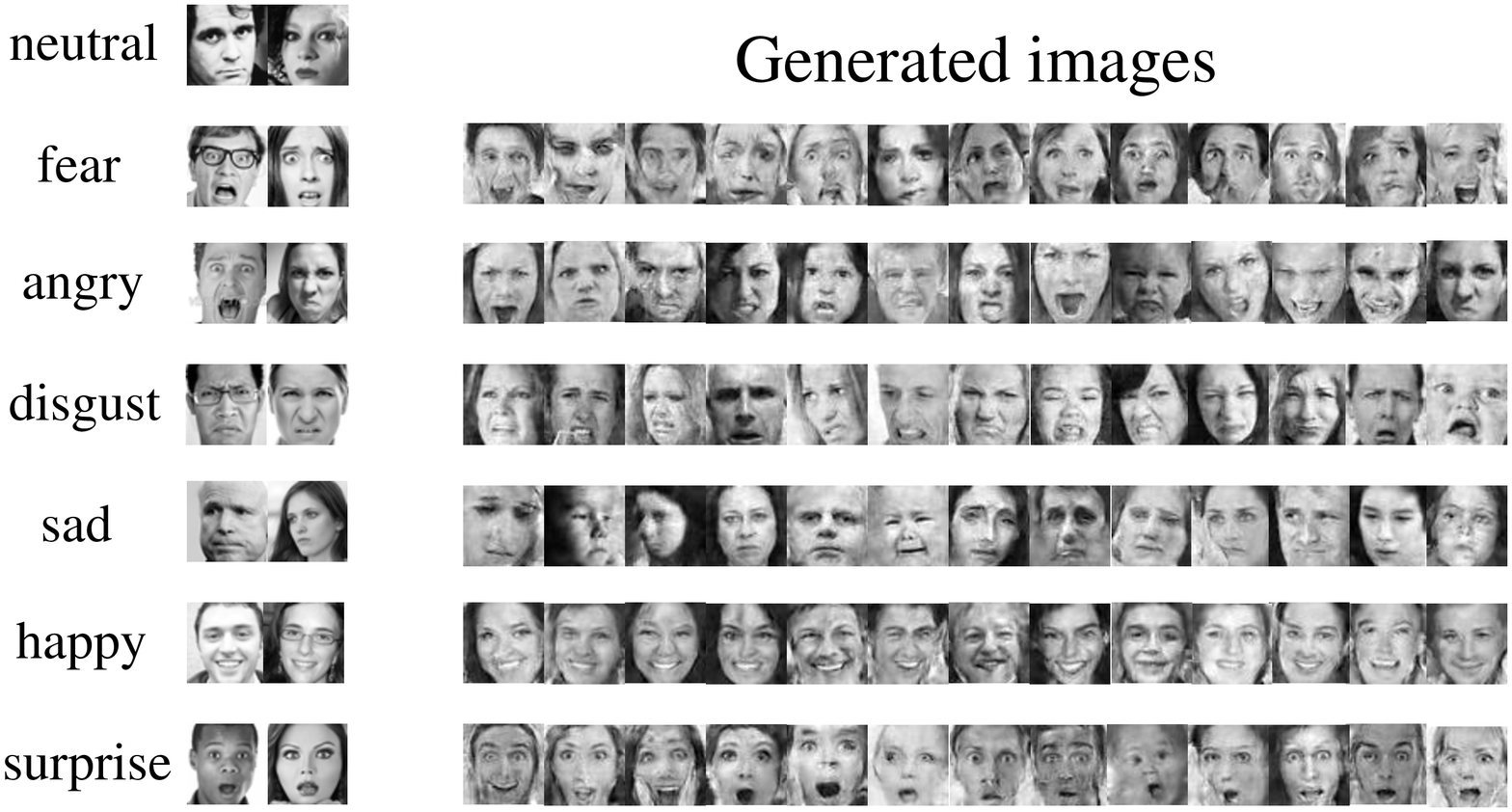}
\par\end{centering}
\par\end{centering}

\caption{\label{fig:face}The original samples and generated samples for each classes. The left two columns are original data and the rest ones are generated by CycleGAN. The neutral class, as reference class, has no generated samples in our experiment.}
\end{figure}

In our experiments, three benchmark datasets are tested:  Facial Expression
Recognition Database (FER2013) \cite{goodfellow2013challenges}, Static Facial Expressions in the Wild (SFEW) \cite{dhall2011static} and Japanese Female Facial Expression (JAFFE) Database \cite{lyons1998japanese}. All these datasets contain 7 types of face emotion including \emph{angry}, \emph{disgust}, \emph{fear}, \emph{happy}, \emph{sad}, \emph{surprise}, and \emph{neutral} (labeled by 0$\sim$6 during training and test process). Samples from FER2013 database are shown in Fig.\ref{fig:face} (left) as an example. The distribution of this dataset is imbalanced. In order to verify the effectiveness of the data augmentation, we sample the images by 20\% for each class in FER2013. We also test our data augmentation model on other two small datasets SFEW and JAFFE. During the training process of CycleGAN, we choose `neutral' class as our reference class and the other six are regarded as target ones, since it is natural to generate faces with emotion from non-emotional ones.\\

\subsection{Experimental Results}
We first train a CNN model based on original FER2013 datasets (20\% sampled) as our baseline and the result is shown in Table \ref{tab:eval1}. We use 7\% and 14\% samples in FER2013 for test.
In order to get the most intuitive results, we choose class $disgust$ and $sad$ from FER2013 as our target classes, which are much smaller than the other classes, as a result, it cannot obtain sufficient learning and optimizing, thus reach a relatively low accuracy when trained on the baseline (see Table \ref{tab:eval1}). In this case, two CycleGANs are trained to generate $disgust$ and $sad$ images respectively with class $neutral$ as reference class (see Fig.\ref{fig:face}), and then are filled into the original datasets (denoted by $+disgust$ and $+sad$) to balance the distribution and complete the data manifold (Table \ref{tab:eval1}).

\begin{table}[!h]
\centering
\caption{\label{tab:eval1} Accuracy of the baseline model (CNN) and our proposed model (CNN+CycleGAN) on FER2013. The best results are highlighted.}
\begin{tabular}{c|c|c c|c|c c}
\hline Class & \multicolumn{3}{|c|}{Accuracy-(7\%)} & \multicolumn{3}{|c}{Accuracy-(14\%)}\\
\hline
 &baseline&+disgust&+sad&baseline&+disgust&+sad\\
\hline
\hline
angry&93.70&\textbf{93.71}&93.05&93.47&93.36&92.89\\
\hline
disgust&73.91&91.30&\textbf{95.65}&79.62&88.89&94.44\\
\hline
fear&90.88&92.18&94.46&90.38&91.43&\textbf{94.58}\\
\hline
happy&91.87&96.34&93.70&91.75&\textbf{96.37}&94.21\\
\hline
sad&87.86&93.61&\textbf{97.44}&89.22&93.26&94.61\\
\hline
surprise&94.27&\textbf{99.12}&96.48&93.46&97.09&96.85\\
\hline
neutral&89.55&91.94&\textbf{94.63}&88.24&93.06&94.48\\
\hline
All&91.04&94.25&\textbf{94.65}&90.77&93.82&94.32 \\
\hline
\par\end{tabular}
\end{table}
From Table \ref{tab:eval1}, we can have the following observations: (a) the overall test accuracy is improved and (b) accuracy of target class raise significantly and it is worth mentioning that (c) the accuracy of reference class $neutral$ also increases. Therefore, we can intuitively verify the capability of CycleGAN in generating reliable images, which is helpful in enlarging minorities. Furthermore, this data augmentation of one class also improves accuracy of other classes, since by generating new samples, the data manifold is further supplemented and becomes more completed, thus make more clearly the margins between classes.

\begin{figure}
\begin{centering}
\begin{centering}
\includegraphics[width=1\textwidth]{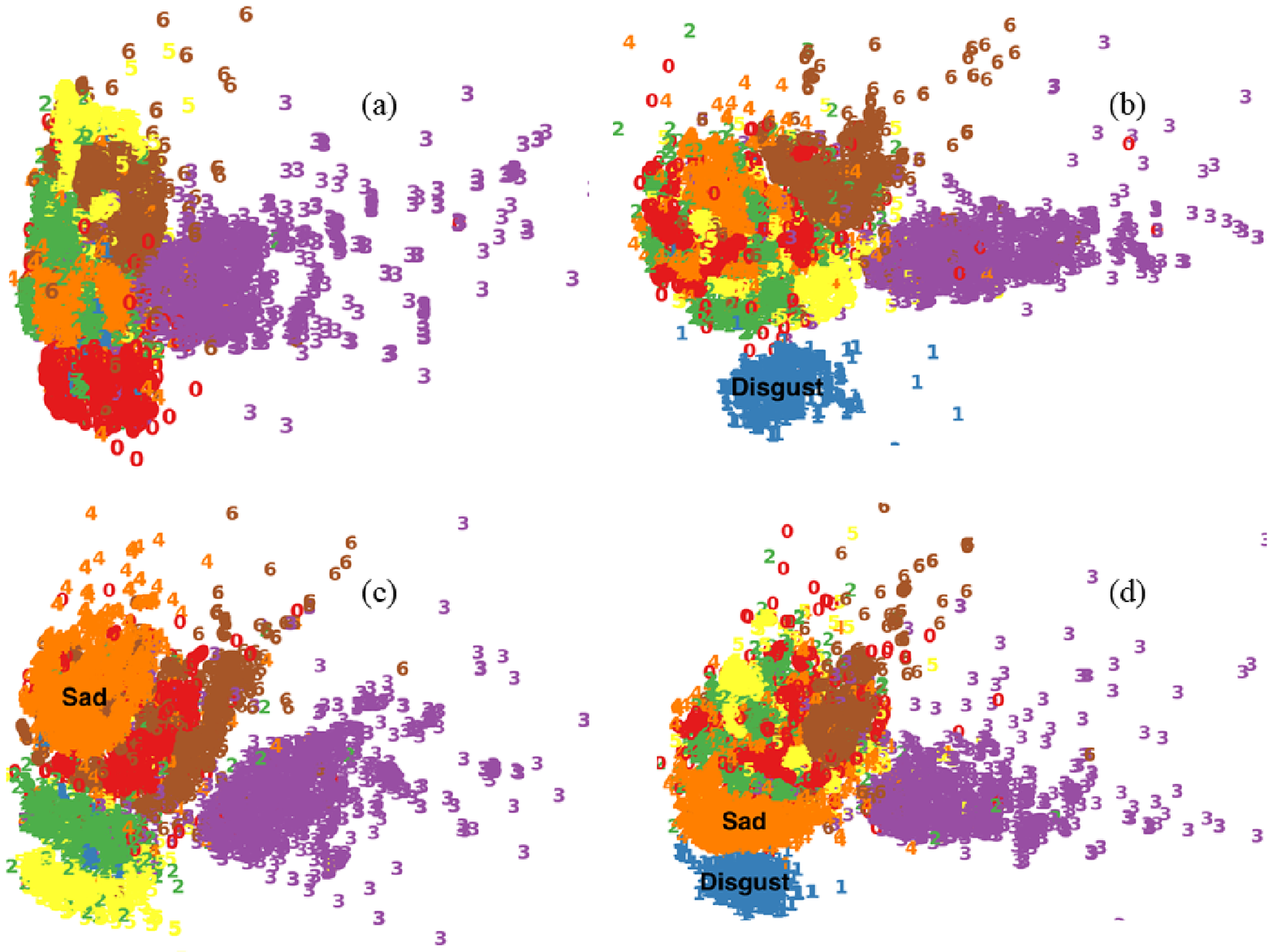}
\par\end{centering}
\par\end{centering}

\caption{\label{fig:manifold_exp}Data manifold of four types of training samples using t-SNE algorithm: (a) baseline model, (b) adding generated $disgust$ samples or (c) $sad$ samples, and (d) samples of both two classes.}
\end{figure}

In order to provide more powerful verification that this data augmentation indeed contributes to the shape of data manifold, we apply a t-distributed stochastic neighbor embedding (t-SNE) algorithm \cite{maaten2008visualizing} to visualize the distribution of training samples by reducing high-dimensional data (48 $\times$ 48) to 2D plane (Fig.\ref{fig:manifold_exp}). Compared to the baseline (Fig.\ref{fig:manifold_exp}-(a)), where sample size of $disgust$ and $sad$ is too small to form a clear margin with other classes, (b) and (c) in Fig.\ref{fig:manifold_exp} shows great improvement in enlarging the sample size, supplementing the data manifold and completing data distribution. Fig.\ref{fig:manifold_exp}-(d) is a much stronger validation where both two classes stand out to improve data manifold.

After generating specific classes to validate GAN's positive role in data augmentation, we make further experiments on our framework based on all three datasets mentioned in Section 4.2. During this process, a baseline model and a model using our data augmentation framework (pre-train+fine-tune) is trained respectively. In our framework, all classes except $neutral$ are generated from CycleGAN and then added as supplementary training corpus for training classification task. (See Fig. \ref{fig:face} for generated images in FER2013 database) and then the model is fine-tuned based on original datasets. Because there is a small number of examples in datasets SFEW and JAFFE, we set the FER2013 database as our pre-trained model and fine-tune it using these two datasets. Similar experiments were reported in \cite{yu2015image}. In order to reduce the inference of complex background in SFEW, we apply a simple cropping method to extract faces from original images. For testing, we use 7\% and 14\% samples from FER2013, the given testing corpus of SFEW and 20\% samples from JAFFE, respectively. Results are shown in Table \ref{tab:all}. In the column DAG: Pre-train + Fine-tune, Pre-train represents the first 10K steps training on generated images from all six classes and Fine-tune represents another 10k fine-tuning steps training on original datasets. SFEW and JAFFE datasets are trained based on the FER2013 model.
\begin{table}
\centering
\caption{\label{tab:all} Test accuracy of baseline and our framework (DAG).}
\begin{tabular}{c|c|c|c|c}
\hline Datasets& \multicolumn{4}{|c}{Accuracy}\\
\hline
&\multicolumn{2}{|c|}{baseline}&\multicolumn{2}{|c}{DAG: pre-train+fine-tune}\\
\hline
\hline
FER2013&91.04(7\%)&90.77(14\%)&\textbf{94.71}(7\%)&94.35(14\%)\\
\hline
FER+SFEW&\multicolumn{2}{|c|}{31.92}&\multicolumn{2}{c}{\textbf{39.07}}\\
\hline
FER+JAFFE&\multicolumn{2}{|c|}{93.87}&\multicolumn{2}{c}{\textbf{95.80}}\\
\hline
\par\end{tabular}

\end{table}
After applying our framework of data augmentation using GAN, accuracy of all the three datasets has visibly improved. As for FER2013 database which has obvious imbalanced distribution among classes, our data augmentation technique is able to complete data manifold, especially for those which have much smaller samples. And for small datasets like SFEW and JAFFE, our technique can generate feature-level synthetic images from existing samples to enlarge the origianl datasets and form clear margins or hyper-planes between neighboring classes.

\section{Conclusions and Discussions}
In this paper, we explored using GAN for data augmentation in task of emotion classification task.
We propose a framework for data augmentation by using CycleGAN to generate auxiliary data to minority classes in training. During the process of training CycleGAN model, a least-squared loss is combined with original adversarial loss to avoid gradient vanishing problem. Besides, we show the GAN's ability of supplementing low-dimensional data manifold. Because of possessing a more complete data manifold, the classifier can be better learned to find margins or hyper-planes of neighboring classes. Experiments on three benchmark datasets show that our GAN-based data augmentation techniques can lead to improvement in distribution integrity and margin clarity between classes, and can obtain 5\%$\sim$10\% increase in the accuracy of emotion classification task.

The work still has some limitations. For instance, the datasets we select are limited to emotions and only CycleGAN is used in our model. Therefore, we consider our future work to apply our model for the general image classification problems, and try other GAN models to evaluate data augmentation method with stronger evidence.
\bibliographystyle{splncs03}
\bibliography{reference}

\appendix
\addtocontents{toc}{\protect\contentsline{chapter}{Appendix:}{}}
\section{Training Details}
\subsection*{CNN Model}
In any stage of classification task, we all apply batch-size = 32, stable learning rate=1e-3 and training step = 20000. Adam optimizer is used whose parameter $\beta{_1}$ = 0.5. More detailed configurations are listed in Table. \ref{tab:cnn}

\begin{table}[htb]
\centering
\begin{tabular}{c|c}
\hline
Layer Type& Configuration\\
\hline
\hline
Input image&48*48*1\\
\hline
Convolution\&ReLU&[3, 3, 1, 64] s=1\\
\hline
Max-Pooling\&Norm&[1, 3, 3, 1] s=2\\
\hline
Convolution\&ReLU&[3, 3, 64, 128] s=1\\
\hline
Max-Pooling\&Norm&[1, 3, 3, 1] s=2\\
\hline
FC*2&256\\
\hline
Softmax&[256, 7]\\
\hline
Output logits&[7]\\
\hline
\end{tabular}\\
\caption{\label{tab:cnn} Configuration of the convolutional neural network, "s" represents stride. FC means fully connected operation and there are two FC layers in this network.}
\par\end{table}

\subsection*{CycleGAN}
During training the CycleGAN model, we use batch-size=1, learning rate=2e-4 and 1e-4. Adam optimizer is used and $\beta{_1}$ is set to 0.5. Besides, the hyper-parameters of CycleGAN are 10 for both $\lambda{_1}$ and $\lambda{_2}$.\\
More detailed configurations are listed in Table.\ref{tab:cycleg} and \ref{tab:cycled}

\begin{table}[htb]
\centering
\begin{tabular}{c|c}
\hline
Layer Type& Configuration\\
\hline
\hline
Input&48*48*1\\
\hline
Conv-BN-ReLU&7*7, 64, s=1\\
\hline
Conv-BN-ReLU&3*3, 128, s=2\\
\hline
Conv-BN-ReLU&3*3, 256, s=2\\
\hline
Res-Block  *6&2 3*3 conv\\
\hline
Deconv-BN-ReLU&3*3, 128, s=1/2\\
\hline
Deconv-BN-ReLU&3*3, 64, s=1/\\
\hline
Conv-BN-ReLU&7*7, 1, s=1\\
\hline
Output&48*48*1\\
\hline
\end{tabular}
\caption{\label{tab:cycleg} Configuration of the generator in CycleGAN, "s" represents stride. Conv, BN, Deconv represent convolution, batch-normalization and deconvolution(matrix transpose) respectively. We apply 6 Resnet block in our network and each block has 2 convolution layers.}
\end{table}
\begin{table}
\centering
\begin{tabular}{c|c}
\hline
Layer Type& Configuration\\
\hline
\hline
Input&48*48*1\\
\hline
Conv-BN-ReLU&4*4, 64, s=2\\
\hline
Conv-BN-ReLU&4*4, 128, s=2\\
\hline
Conv-BN-ReLU&4*4, 256, s=2\\
\hline
Conv-BN-ReLU&4*4, 512, s=2\\
\hline
Conv-BN-ReLU&4*4, 1, s=1\\
\hline
Output&1\\
\hline
\end{tabular}
\caption{\label{tab:cycled} Configuration of the discriminator in CycleGAN, "s" represents stride. Settings and representations are same as generator.}
\end{table}

\end{document}